\documentclass[10pt,twocolumn,letterpaper]{article}

\usepackage{3dv}
\usepackage{times}
\usepackage{epsfig}
\usepackage{graphicx}
\usepackage{amsmath}
\usepackage{amssymb}

\usepackage{enumitem}
\usepackage{array}
\usepackage{cite}
\usepackage{stmaryrd}
\usepackage{multirow}
\usepackage{algorithm}
\usepackage{algorithmic}
\usepackage[T1]{fontenc}
\usepackage{multicol}
\usepackage{stfloats}
\usepackage[caption = false]{subfig}
\newcommand\mydots{\hbox to 1em{.\hss.\hss.}}

\usepackage[pagebackref=true,breaklinks=true,letterpaper=true,colorlinks,bookmarks=false]{hyperref}

\threedvfinalcopy 

\def\threedvPaperID{51} 

\ifthreedvfinal\fi
\begin{document}

\title{Semantic Classification of 3D Point Clouds with Multiscale Spherical Neighborhoods}

\author{
Hugues Thomas
\quad
Jean-Emmanuel Deschaud
\quad
Beatriz Marcotegui
\\
Fran\c{c}ois Goulette
\quad
Yann Le Gall
\\
\\
Mines ParisTech, PSL Research University\\
{\tt\small firstname.lastname@mines-paristech.fr}
}

\maketitle

\begin{abstract}
This paper introduces a new definition of multiscale neighborhoods in 3D point clouds. This definition, based on spherical neighborhoods and proportional subsampling, allows the computation of features with a consistent geometrical meaning, which is not the case when using k-nearest neighbors. With an appropriate learning strategy, the proposed features can be used in a random forest to classify 3D points. In this semantic classification task, we show that our multiscale features outperform state-of-the-art features using the same experimental conditions. Furthermore, their classification power competes with more elaborate classification approaches including Deep Learning methods.
\end{abstract}

\section{Introduction}

\begin{figure*}[b!]
    \centering
    \includegraphics[width=\textwidth, keepaspectratio=true]{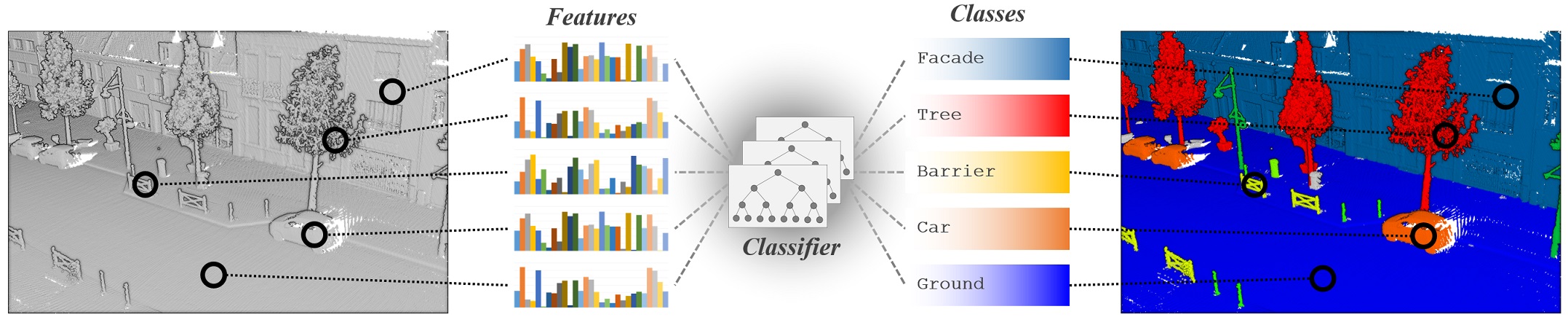}
    \centering
    \caption{Our 3D semantic classification framework: a set of features is computed for every point with our new multiscale spherical neighborhood definition. These features are then used to classify each point independently.}
    \label{fig_intro}
\end{figure*}

In the past years, the interest in 3D scanning technologies has constantly grown in the computer vision community. The benefits of combining 3D and semantic information are fundamental for robotic applications or autonomous driving. To assign the right label to every point in a 3D scene, semantic classification algorithms need to understand the geometry of the scene. Among the ways to achieve such an understanding, two paradigms stand out. In the first instance, the point cloud is segmented and then a label is given to each segment \cite{golovinskiy_shape-based_2009, serna_detection_2014, roynard_fast_2016}. The weakness of this first strategy is that it depends on a prior segmentation which does not use semantic information. In the second instance, each point is considered individually and is given a semantic label or class probabilities (see Figure \ref{fig_intro}) \cite{weinmann_semantic_2015, hackel_fast_2016}. Without any prior segmentation, classifying 3D points only relies on the appearance of points neighborhoods. Thus, we need a set of expressive features to describe the geometry in a point neighborhood. Demantk\'e et al. proposed a description based on local covariance \cite{demantke_dimensionality_2011}. To complement this local shape description, Weinmann et al. added measures of verticality and height distribution \cite{weinmann_feature_2013}. We could find more complex descriptors in the literature like \textit{Spin Images} \cite{johnson_using_1999} or \textit{Fast Point Feature Histograms} \cite{rusu2009fast}. However, we chose to use a multiscale approach with simple features, which has been proven to be more expressive \cite{hackel_fast_2016}. 

Standard machine learning techniques are used to classify 3D points described by geometric features. According to Weinmann et al.'s extensive work \cite{weinmann_semantic_2015}, \textit{Random Forest} is the most suitable classifier. In that case, spatial relations between points are ignored. Features can instead be used as unary potentials in \textit{Markov Random Fields} \cite{munoz_onboard_2009} to ensure spatial coherence in the classification. These two techniques can also be combined by using class probabilities given by a standard classifier as unary potentials \cite{shapovalov2010nonassociative}. In that case, the random fields can be seen as a subsequent semantic segmentation, and the first point-wise classification problem remains. We focus on this first classification, because the better its results are, the better further processing will perform.

3D neural networks have recently been used for 3D Semantic Segmentation. Huang et al. \cite{huang_point_2016} used a 3D version of fully convolutional neural networks (FCNN) to label point clouds from voxel-wise predictions. An original Multilayer Perceptron (MLP) architecture, named Pointnet, has been presented by Qi et al. \cite{qi_pointnet_2017}, and is able to extract both global and local features from a 3D point cloud. Its extension, Pointnet++ \cite{qi2017pointnet++}, achieves even better results by the aggregation of local features in a hierarchical manner. 3D neural networks can also be combined with graph-based segmentation methods to design more elaborate 3D semantic segmentation algorithms. Tchapmi et al. \cite{tchapmi_segcloud_3dv17} used a CRF on 3D FCNN predictions to enforce global consistency and provide fine-grained semantics. On the other hand, Landrieu and Simonovsky \cite{landrieu_large-scale_2017} prefer to segment the cloud in a Superpoint Graph first, and then use Pointnet architecture and graph convolutions to classify each Superpoint. Even though hand-crafted features rarely perform at the level of Deep Learning architectures, our multiscale features compete with most of these methods.

Although we propose a slightly different set of features than Hackel et al. \cite{hackel_fast_2016}, the originality of our method lies in the points neighborhood selection. In the case of 3D points classification, the two most commonly used neighborhood definitions are the \textit{spherical neighborhood} \cite{brodu_3d_2012} and the \textit{k-nearest neighbors} (KNN) \cite{weinmann_feature_2013}. For a given point $P$, the spherical neighborhood comprises the points situated less than a fixed radius from $P$, and the k-nearest neighbors comprises a fixed number of closest points to $P$. We can also add a third definition, which is mostly used for airborne lidar data \cite{chehata2009airborne, niemeyer2014contextual}, the \textit{cylindrical neighborhood} which comprises the points situated less than a fixed radius from $P$, on a 2D projection of the cloud (frequently on the horizontal plane). Whatever definition is chosen, the scale of the neighborhood has to be determined. Using a fixed scale across the scene is inadequate because most scenes contain objects of various sizes. Weinmann et al. explored a way to adapt the scale to each point of the cloud \cite{weinmann_semantic_2015}. However, using a multiscale approach has proven to be more effective, whether it is used with KNN \cite{pauly2003multi, hackel_fast_2016}, with spherical/cylindrical neighborhoods \cite{brodu_3d_2012, niemeyer2014contextual} or with a combination of all neighborhood types \cite{blomley2016classification}. The major drawback of the multiscale neighborhoods is their computational time, but Hackel et al. \cite{hackel_fast_2016} suggested a simple and efficient solution to implement them, based on iterative subsamplings of the cloud. However, their definition of multiscale neighborhoods, using KNN, lacks geometrical meaning. Section 2. describes our definition of \textbf{multiscale spherical neighborhoods}, which keeps the features undistorted while ensuring sufficient density at each scale.

We chose to evaluate our multiscale spherical neighborhoods definition on a semantic classification basis. The features we use and our learning strategy are described in Section 3. We conduct several experiments detailed in Section 4 on various datasets. First, we validate that our multiscale features outperform state of the art features in the same experimental conditions on two small outdoor datasets. Then we compare our classification results to more elaborate semantic segmentation methods on three bigger datasets. The parameters' influence is eventually highlighted in the last paragraph.


\section{Multiscale Spherical Neighborhoods}

\begin{figure}[b]
    \centering
    \includegraphics[width=\columnwidth, keepaspectratio=true]{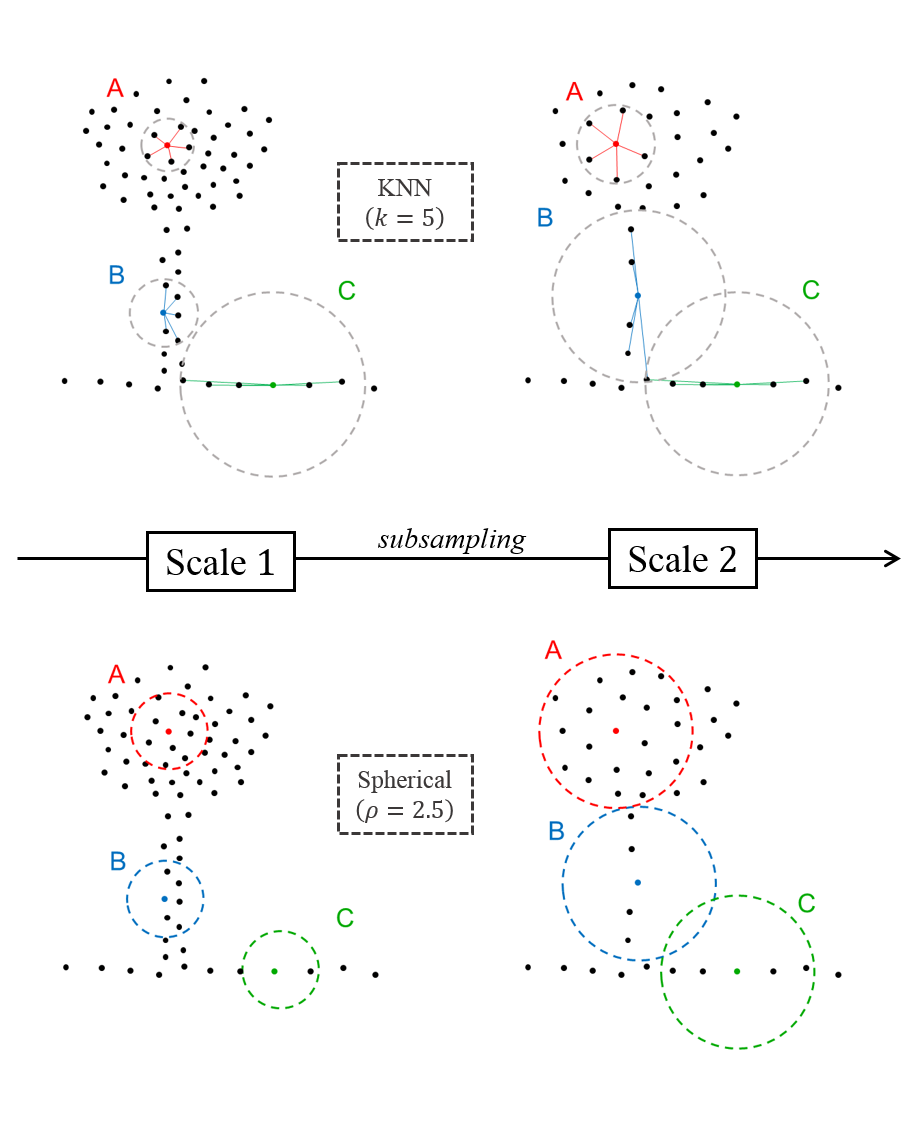}
    \centering
    \caption{Behavior of multiscale neighborhoods defined with KNN or with spherical neighborhoods.}
    \label{fig_multiscale}
\end{figure}

Our new definition of multiscale neighborhoods is inspired by \cite{hackel_fast_2016} with spherical neighborhoods instead of KNN. This section highlights the differences between both definitions. Let $\mathcal{C}\subset\mathbb{R}^3$ be a point cloud, the spherical neighborhood of point $\mathbf{p}_0\in\mathbb{R}^3$ in $\mathcal{C}$ with radius $r \in \mathbb{R}$ is defined by:
\begin{equation}
    \mathfrak{S}_r(\mathbf{p}_0,\mathcal{C}) 
    = 
    \left\lbrace
        \begin{array}{c|c}  
            \mathbf{p}\in\mathcal{C} &  \left\Vert \mathbf{p}-\mathbf{p}_0 \right\Vert \leqslant r 
        \end{array} 
    \right\rbrace
\end{equation}
Unlike KNN, this neighborhood corresponds to a fixed part of the space as shown in Figure \ref{fig_multiscale}. This property is the key to give a more consistent geometrical meaning to the features. But, in that fixed part, the number of points can vary according to the cloud density. As Hackel et al. \cite{hackel_fast_2016} explained, radius search should be the correct procedure from a purely conceptual viewpoint but it is impractical if the point density exhibits strong variations. Two phenomena appear in particular:
\begin{itemize}
  \item Having too many points when the neighborhood scale is too big or the density too high
  \item Having too few points when the neighborhood scale is too small or the density too low
\end{itemize}
The first phenomenon has computational consequences as getting a large number neighbors in a larger set of points takes a lot of time. The computational cost of multiscale features does not come from the fact that we have to compute the features $S$ times, where $S$ is the number of scales. The real limiting factor is the number of points contained in the biggest scales. Furthermore, all those points are not required to compute relevant features. Our features capture a global shape in the neighborhood and do not need fine details. The solution proposed by Hackel et al. \cite{hackel_fast_2016} to subsample the cloud proportionally to the scale of the neighborhood can be adapted to the spherical definition. This solution better suits the spherical neighborhoods than the KNN. With a uniform density, the number of points in the neighborhood becomes a feature itself, describing the neighborhood occupancy rate. We chose to subsample the cloud with a grid, by keeping the barycenter of the points comprised in each cell. Let $l \in \mathbb{R}$ be the size of the grid cells, for any radius of a neighborhood $r \in \mathbb{R}$, we can control the maximum number of points in our neighborhood with the parameter $\rho=\frac{r}{l}$. If $\rho$ is too low, we will not have enough points and the features will not be discriminant, but the higher its value is, the longer computations are.

The impact of the second phenomenon, caused by low densities, should be limited by the use of multiple scales. As illustrated in Figure \ref{fig_multiscale}, if the density is too low, the points will remain the same after subsampling between two consecutive scales (neighborhood C). With spherical neighborhoods, the small scale might not contain enough points for a good description, but the large scale will deliver the information. In the same case, the KNN behave differently, giving exactly the same information at both scales. Regardless of the neighborhoods, there is no information to get from the data at the smaller scale. However, the KNN give a false description of the smaller scale without any measure of its reliability, whereas spherical neighbors give the number of points, which is an indication of the robustness of the description.

The scales are defined by three parameters: the radius of the smallest neighborhood $r_0$, the number of scales $S$, and the ratio between the radius of consecutive neighborhoods $\varphi$. We can then define the neighborhood at each scale $s \in \left\lbrace 0,...,S-1 \right\rbrace$ around point $\mathbf{p}_0\in\mathbb{R}^3$ as: 
\begin{equation}
    \mathcal{N}_s(\mathbf{p}_0) = \mathfrak{S}_{r_s}(\mathbf{p}_0,\mathcal{C}_s) 
\end{equation}
with $r_s = r_0 * \varphi ^ s$ being the radius at scale $s$ and $\mathcal{C}_s$ being the cloud subsampled with a grid size of $r_s/\rho$. 

Despite its similarity with the definition proposed by Hackel et al. \cite{hackel_fast_2016}, our multiscale neighborhood definition stands out with its geometrical meaning. With spherical neighborhoods instead of KNN, the features always describe a part of the space of the same size at each scale. Moreover, the number of points in the neighborhood is now a feature itself adding even more value to this definition. More than a theoretical good behaviour, this leads to better feature performances, as shown in Section 4.


\section{Point-wise Semantic Classification}

\subsection{Geometric and Color Features}
\begin{table}[b]
\centering
\begin{small}
\caption{Features used for classification.}
\label{Table_Features} 
\vspace{2ex}
\begin{tabular}{| r | c |}
    \hline 
    \multicolumn{1}{|c|}{Features} & Definitions \rule{-3pt}{3.2 ex} \\[0.8 ex]
    \hline 
    Sum of eigenvalues & \(\displaystyle \sum\lambda_i\) \rule{-3pt}{3.2 ex} \\[0.8 ex]
    Omnivariance & \(\displaystyle \left(\prod\lambda_i\right)^{\frac{1}{3}}\) \\[2ex]
    Eigenentropy & \(\displaystyle -\sum\lambda_i \ln(\lambda_i)\) \\[2 ex]
    Linearity & \(\displaystyle (\lambda_1 - \lambda_2) / \lambda_1\) \\[2ex]
    Planarity & \(\displaystyle (\lambda_2 - \lambda_3) / \lambda_1\) \\[2 ex]
    Sphericity & \(\displaystyle \lambda_3 / \lambda_1\) \\[2ex]
    Change of curvature & \(\displaystyle \lambda_3 / (\lambda_1 + \lambda_2 + \lambda_3)\) \\[2 ex]
    Verticality (x2) & \(\displaystyle \left|\frac{\pi}{2} - \mathsf{angle}(\mathbf{e}_i, \mathbf{e}_z)\right|_{i \in (0, 2)}\) \\[2 ex]
    Absolute moment (x6) & \(\displaystyle \frac{1}{|\mathcal{N}|}{\left|\sum\langle \mathbf{p}-\mathbf{p}_0 , \mathbf{e}_i \rangle^k\right|}_{i \in (0, 1, 2)} \) \\[2 ex]
    Vertical moment (x2) & \(\displaystyle \frac{1}{|\mathcal{N}|}\sum\langle \mathbf{p}-\mathbf{p}_0 , \mathbf{e}_z \rangle^k \) \\[2 ex]
    Number of points & \(\displaystyle |\mathcal{N}|\) \\[1 ex]
    \hline 
    Average color (x3) & \(\displaystyle \frac{1}{|\mathcal{N}|}\sum c\) \rule{-3pt}{3.8 ex} \\[2 ex]
    Color variance (x3) & \(\displaystyle \frac{1}{|\mathcal{N}|-1}\sum (c - \bar{c})^2 \) \\[2 ex]
    \hline 
\end{tabular}
\end{small}
\end{table}

For benchmarking purposes, we divide our features in two sets described in Table \ref{Table_Features}. The first set does not use any additional information like intensity, color, or multispectral measure, to keep previous work conditions \cite{weinmann_semantic_2015, hackel_fast_2016} in our first experiment (Section 4.1). In the other experiments, additional color features are used when available. We use covariance based features that simply derive from the eigenvalues $\lambda_1 \geqslant \lambda_2 \geqslant \lambda_3 \in \mathbb{R}$ and corresponding eigenvectors $\mathbf{e}_1, \mathbf{e}_2, \mathbf{e}_3 \in \mathbb{R}^3$ of the neighborhood covariance matrix defined by:
\begin{equation}
    \mathsf{cov}\!\left(\mathcal{N}\right)=\frac{1}{|\mathcal{N}|}\sum_{\mathbf{p}\in\mathcal{N}}(\mathbf{p}-\bar{\mathbf{p}})(\mathbf{p}-\bar{\mathbf{p}})^T
\end{equation}
Where $\bar{\mathbf{p}}$ is the centroid of the neighborhood $\mathcal{N}$. From the eigenvalues, we can compute several features: \textit{sum of eigenvalues}, \textit{omnivariance}, \textit{eigenentropy}, \textit{linearity}, \textit{planarity}, \textit{sphericity}, \textit{anisotropy}, and \textit{change of curvature}. However, among those commonly used features, we eliminate \textit{anisotropy} defined by $(\lambda_1 - \lambda_3) / \lambda_1$ as it is strictly equivalent to \textit{sphericity}. We can notice that, thanks to the nature of our neighborhoods, we do not need to normalize the eigenvalues as in previous works. Their values do not vary with the original point cloud density which means the features that do not involve ratios, e. g. \textit{sum of eigenvalues}, \textit{omnivariance}, and \textit{eigenentropy}, make more sense. 
Our feature set is completed by \textit{verticality} that we redefined as $|\frac{\pi}{2} - \mathsf{angle}(\mathbf{e}_i, \mathbf{e}_z)|$. Unlike Hackel et al. \cite{hackel_fast_2016}, we keep the verticality for the first and the last eigenvectors. The first one encodes the verticality of linear objects, and the last one the verticality of the normal vector of planar objects. We also use first and second order \textit{moments} around all three eigenvectors, but in absolute value as the eigenvectors have random orientations. Following our assumption that vertical direction plays an important role, additional \textit{vertical moments} are computed around the vertical vector $\mathbf{e}_z$ in relative value as the upward direction is always the same. Eventually, as explained in Section 2, the number of points in a neighborhood completes our first set of features, which contains $18$ values at each scale.
In Section 4.2, we use colors as previous works did, because some objects like closed doors or windows are indistinguishable in 3D. We chose simple features, the mean and the variance of each color channel, bringing the total number of features per scale to $24$.

\subsection{Learning Strategy}

There are two setbacks when classifying a point cloud. First, its size is generally huge, and then, the classes are heavily unbalanced. To fix those problems, one can take a subset of the training data, small enough to allow reasonable training times, and balance the classes in that subset. The scope of the results also depends on the test set. With small datasets, the rest of the points are used as the test set, even if they represent the same scene. The results from such experiments would be questionable as a measure of the classification performances, however, they still may be used to compare the descriptive power of different features. The recent appearance of bigger point cloud datasets allowed the separation of the training set and the test set. With such point clouds, it is possible to get a relevant measure of how well the classification generalizes to unseen data. 

In Section 4.1, we compare our multiscale features to state of the art features \cite{weinmann_semantic_2015, hackel_fast_2016} in the same experimental conditions. The same number of points is randomly picked in each class to train a classifier, and this classifier is tested on the rest of the cloud. We go further than previous works by computing our results several times with different training sets. We cannot ensure that the comparison is valid without checking the distribution of results on a large number of trials. The quality of our multiscale features can be assessed more reliably in these conditions.

On bigger datasets, we use a different learning strategy. We iteratively add points to the training set with a trial and error procedure. A classifier is trained on a set of points $\mathcal{T}$ from the training clouds $\mathcal{U}$, then the classifier is tested on $\mathcal{U}$, and we randomly add some of the misclassified points to $\mathcal{T}$. After some iterations, the classifier is used on the test clouds. The experiments in Section 4.2 use this learning strategy, which only consists in a smart choice of the training points. We can't use this strategy on small datasets, because the test scene is the same as the training scene and our classification would show overfitted results.


\section{Experiments}

\subsection{State of the art features comparison}

\newcolumntype{C}[1]{>{\centering\let\newline\\\arraybackslash\hspace{0pt}}p{#1}}
\newcolumntype{R}[1]{>{\raggedleft\let\newline\\\arraybackslash\hspace{0pt}}p{#1}}
\newcommand*\rot{\rotatebox{90}}
\newcommand\Tstrut{\rule{-3pt}{2.6ex}}       
\newcommand\Bstrut{\rule[-0.9ex]{-3pt}{0pt}} 
\newcommand{\TBstrut}{\rule{-3pt}{2.6ex} \rule[-0.9ex]{-2pt}{0pt}}  

\begin{table}[b]
\centering
\caption{Average $\mathsf{IoU}$ (with standard deviation) on Rue Madame (top) and Rue Cassette (bottom) datasets. Results for \cite{hackel_fast_2016} and \cite{weinmann_semantic_2015} are converted from corresponding articles.}
\vspace{1ex}
\label{Table_SoA} 
\begin{small}
\begin{tabular}{| C{1.65cm} | C{2.4cm} | C{1.2cm} | C{1.2cm} |}   
    \multicolumn{1}{c}{Class} & \multicolumn{1}{c}{Ours} & \multicolumn{1}{c}{\cite{hackel_fast_2016}} & \multicolumn{1}{c}{\cite{weinmann_semantic_2015}} \TBstrut\\
    \hline
    Facade &	$\mathbf{98.22\%}\,(\pm0.11)	$ & $97.06\%$ &	$91.81\%$ \Tstrut\\
    Ground	&	$\mathbf{96.62\%}\,(\pm0.18)	$ & $96.29\%$ &	$84.88\%$ \\
    Cars	&	$\mathbf{95.37\%}\,(\pm0.43)	$ & $89.09\%$ &	$55.48\%$ \\
    Motorcycles	&	$\mathbf{61.55\%}\,(\pm2.22)	$ & $47.44\%$ &	$9.44\%$ \\
    Traffic Signs	&	$\mathbf{67.43\%}\,(\pm8.67)	$ & $33.96\%$ &	$4.90\%$ \\
    Pedestrians	&	$\mathbf{77.86\%}\,(\pm3.32)	$ & $24.13\%$ &	$1.63\%$ \Bstrut\\
    \hline
    \textbf{Mean}	&	$\mathbf{82.84\%} 		$ & $58.89\%$ &	$31.68\%$ \TBstrut\\
    \hline 
\end{tabular}
\end{small}

\vspace{0.05cm}
\begin{small}
\begin{tabular}{| C{1.65cm} | C{2.4cm} | C{1.2cm} | C{1.2cm} |}    
    \hline
    Facade	&	$\mathbf{97.27\%}\,(\pm0.20)	$ & $93.89\%$ &	$86.65\%$ \Tstrut\\
    Ground	&	$\mathbf{97.77\%}\,(\pm0.20)	$ & $96.99\%$ &	$95.75\%$ \\
    Cars	&	$\mathbf{84.94\%}\,(\pm1.54)	$ & $80.88\%$ &	$47.31\%$ \\
    Motorcycles	&	$\mathbf{58.99\%}\,(\pm3.27)	$ & $51.33\%$ &	$17.12\%$ \\
    Traffic Signs	&	$12.71\%\,(\pm3.81)	$ & $\mathbf{18.58\%}$ &	$14.29\%$ \\
    Pedestrians	&	$\mathbf{35.31\%}\,(\pm3.80)	$ & $24.69\%$ &	$9.06\%$ \\
    Vegetation	&	$\mathbf{71.48\%}\,(\pm1.94)	$ & $51.40\%$ &	$24.63\%$ \Bstrut\\
    \hline
    \textbf{Mean}	&	$\mathbf{65.50\%} 		$ & $54.08\%$ &	$35.30\%$ \TBstrut\\
    \hline
\end{tabular}
\end{small}
\end{table}

The goal of our first experiment is to assess the performances of our multiscale features against other state-of-the-art features \cite{weinmann_semantic_2015, hackel_fast_2016}, thus, we keep the same experimental conditions as Weinmann et al. \cite{weinmann_semantic_2015} and Hackel et al. \cite{hackel_fast_2016}. We use the Paris-Rue-Madame dataset \cite{serna_paris-rue-madame_2014}, a 160-meter street scan containing 20 million points and the Paris-Rue-Cassette dataset \cite{vallet2015terramobilita}, a 200-meter street scan containing 12 million points. To focus the comparison on the features, we use a random forest classifier trained on 1000 random points per class for each dataset as previous works did. We chose the parameters $S=8$, $r_0=0.1m$, $\varphi=2$ and $\rho=5$. The first three parameters were chosen so that the scales of our neighborhoods cover a range from the smallest object size to the order of magnitude of a facade and the last parameter $\rho$ was chosen empirically (see Section 4.3). With an average personal computer setup (32 GB RAM, Intel Core i7-3770; 3.4 GHz), our feature extraction took about 319 seconds on Rue Cassette, which is the same order of magnitude as the method in \cite{hackel_fast_2016} (191s) and way faster than the method in \cite{weinmann_semantic_2015} (23000s). 

For consistency, we chose to use the classes "Intersection over Union" metric in all our experiments. Thus, we convert the class $F_1$ scores given in \cite{weinmann_semantic_2015, hackel_fast_2016} with the equation :

\begin{equation}
    F_1 = \frac{2\mathsf{TP}}{2\mathsf{TP} + \mathsf{FP} + \mathsf{FN}}
\end{equation}
\begin{equation}
    \mathsf{IoU} = \frac{\mathsf{TP}}{\mathsf{TP} + \mathsf{FP} + \mathsf{FN}} = \frac{F_1}{2 - F_1}
\end{equation}

where $\mathsf{TP}$, $\mathsf{FP}$, and $\mathsf{FN}$ respectively denote true positives, false positives, and false negatives for each class. As stated in Section 3.2, we reproduce our results 500 times to ensure the validity of the comparison despite the random factor in the choice of the training set. In Table \ref{Table_SoA}, we report the average class $\mathsf{IoU}$ to compare with previous results and the standard deviations to prove the consistency of the classification. The performances of our multiscale features exceed previous results by $24$ mean $\mathsf{IoU}$ points on Rue Madame and $11$ mean $\mathsf{IoU}$ points on Rue Cassette. We can also note that our results do not vary much, the standard deviations being limited to a few percents even for the hardest classes with fewer points. 

\begin{table*}[b]
\centering
\begin{footnotesize}
\caption{$\mathsf{IoU}$ per class on S3DIS dataset}
\vspace{2ex}
\label{Table_S3DIS} 
\begin{tabular}{ c | c  c  c  c  c  c  c  c  c  c  c  c | c }

Method	 & ceiling	 & floor	 & wall	 & beam	 & column	 & window	 & door	 & chair	 & table	 & bookcase	 & sofa	 & board	 & mean	\Bstrut\\
\hline
PointNet \cite{qi_pointnet_2017} \footnotemark[1] & $88.8$	& $\mathbf{97.3}$	& $69.8$	& $\mathbf{0.1}$	& $3.9$	& $46.3$	& $10.8$	& $52.6$	& $58.9$	& $40.3$	& $5.9$	& $\mathbf{26.4}$	& $41.7$	\Tstrut\\
SEGCloud \cite{tchapmi_segcloud_3dv17}	& $90.1$	& $96.1$	& $69.9$	& $0.0$	& $18.4$	& $38.4$	& $23.1$	& $75.9$	& $70.4$	& $58.4$	& $40.9$	& $13.0$	& $49.5$	\\
SPGraph \cite{landrieu_large-scale_2017}	& $89.4$	& $96.9$	& $\mathbf{78.1}$	& $0.0$	& $\mathbf{42.8}$	& $\mathbf{48.9}$	& $\mathbf{61.6}$	& $\mathbf{84.7}$	& $\mathbf{75.4}$	& $\mathbf{69.8}$	& $\mathbf{52.6}$	& $2.1$	& $\mathbf{58.5}$	\\
\hline
RF\_MSSF (ours) 	& $\mathbf{95.9}$	& $96.4$	& $67.6$	& $0.0$	& $11.9$	& $48.3$	& $28.8$	& $64.4$	& $68.9$	& $58.6$	& $33.9$	& $22.3$	& $49.8$	\Tstrut
\end{tabular}
\end{footnotesize}
\end{table*}

As a conclusion, the low standard deviation validates the random selection of the training set and legitimates the comparison of the different sets of features. Our multiscale features thus proved to be superior to state of the art features. The difference between Hackel et al.'s multiscale features \cite{hackel_fast_2016} and ours may seem like an implementation detail, with radius neighborhoods instead of KNN. However, the results prove that the type of local neighborhood definition has a great impact on the robustness of the features.

\subsection{Results on large scale data}

Our second experiment shows how our classification method generalizes to unseen data. As shown in Table \ref{Table_datasets}, we chose three large scale datasets from different environments and acquisition methods. With these datasets, we use the smart choice of training points described in Section 3.2 and a random forest classifier. This simple classification algorithm is designed to focus on the point-wise descriptive power of our features, we called it RF\_MSSF for "Random Forest with Multi-Scale Spherical Features."

\begin{table}[H]
\centering
\caption{Dataset characteristics}
\vspace{2ex}
\label{Table_datasets}
\begin{small}
\begin{tabular}{c|c|c|c}

    Name & S3DIS & Semantic3D & Paris-Lille-3D \TBstrut\\
    \hline
    Environment & Indoor & Outdoor & Outdoor \Tstrut\\
    Acquisition & Cameras & Fixed lidar & Mobile lidar \\
    Colors & Yes & Yes & No \\
    Points & $273M$  & $4009M$ & $140M$ \\
    Covered area & $6000\,m^2$ & $110000\,m^2$ & $55000\,m^2$ \Bstrut\\
    
\end{tabular}
\end{small}
\end{table}

\begin{table*}[t]
\centering
\begin{footnotesize}
\caption{$\mathsf{IoU}$ per class on Semantic3D dataset (Fold 5)}
\vspace{2ex}
\label{Table_Semantic3D} 
\begin{tabular}{ c | c  c  c  c  c  c  c  c | c }
\multirow{2}{*}{Method}	& man-made	& natural	& high	& low	& \multirow{2}{*}{buildings}	& hard	& scanning	& \multirow{2}{*}{cars}	& \multirow{2}{*}{mean}	\\
	& terrain	& terrain	& vegetation	& vegetation	&	& scape	& artefacts	& 	& 	\Bstrut\\
\hline
TMLC-MSR \cite{hackel_fast_2016} 	& $89.8$	& $74.5$	& $53.7$	& $26.8$	& $88.8$	& $18.9$	& $36.4$	& $44.7$	& $54.2$	\Tstrut\\
DeePr3SS \cite{lawin2017deep} 	& $85.6$	& $83.2$	& $74.2$	& $32.4$	& $89.7$	& $18.5$	& $25.1$	& $59.2$	& $58.5$	\\
SnapNet \cite{boulch_unstructured_2017} 	& $82.0$	& $77.3$	& $79.7$	& $22.9$	& $91.1$	& $18.4$	& $37.3$	& $64.4$	& $59.1$	\\
SEGCloud \cite{tchapmi_segcloud_3dv17} 	& $83.9$	& $66.0$	& $86.0$	& $40.5$	& $91.1$	& $30.9$	& $27.5$	& $64.3$	& $61.3$	\\
SPGraph \cite{landrieu_large-scale_2017} 	& $\mathbf{97.4}$	& $\mathbf{92.6}$	& $\mathbf{87.9}$	& $\mathbf{44.0}$	& $\mathbf{93.2}$	& $\mathbf{31.0}$	& $\mathbf{63.5}$	& $\mathbf{76.2}$	& $\mathbf{73.2}$	\\
\hline
RF\_MSSF (ours) 	& $87.6$	& $80.3$	& $81.8$	& $36.4$	& $92.2$	& $24.1$	& $42.6$	& $56.6$	& $62.7$	\Tstrut
\end{tabular}
\end{footnotesize}
\end{table*}

Stanford Large-Scale 3D Indoor Spaces Dataset (S3DIS) \cite{armeni_cvpr16} was acquired by 3D cameras and covers six large-scale indoor areas from three different buildings for a total of $6000\,m^2$ and 273 million points.\footnotetext[1]{As Pointnet was evaluated in a k-fold strategy in the original paper, we obtained the results on this particular split from the authors.} To keep the same experimental conditions as Tchapmi et al. \cite{tchapmi_segcloud_3dv17}, we use the fifth area as the test set and train on the rest of the data. Original annotation comprises 12 semantic elements which pertain to the categories of structural building elements (\textit{ceiling}, \textit{floor}, \textit{wall}, \textit{beam}, \textit{column}, \textit{window}, and \textit{door}) and commonly found furniture (\textit{table}, \textit{chair}, \textit{sofa}, \textit{bookcase}, and \textit{board}). A \textit{clutter} class exists as well for all other elements. This last class has no semantic meaning like the "unclassified" points in the other datasets and will not be considered during training and testing. As the object scales in this dataset are smaller than the object scales in a street, we adapt the parameters to $S=8$, $r_0=0.05m$, $\varphi=2$, and $\rho=5$. The classifier is trained on 50000 sample points chosen with the procedure described in Section 3.2. Table \ref{Table_S3DIS} shows that our classification method outperforms the deep learning architectures of \cite{qi_pointnet_2017, tchapmi_segcloud_3dv17} but is unable to compete with the cutting edge algorithm of Landrieu and Simonovsky \cite{landrieu_large-scale_2017}. 

Semantic3D \cite{hackel_semantic3d._2017} is an online benchmark comprising several fixed lidar scans of different outdoor places. This is currently the dataset with the highest number of points (more than 4 billion), and the greatest covered area (around $110000\,m^2$). We kept the parameters used in previous outdoor experiment: $S=8$, $r_0=0.1m$, $\varphi=2$, and $\rho=5$. Table \ref{Table_Semantic3D} provides our results on the \textit{reduced-8} challenge. Our classification method ranked second at the time of the submission. Once again, it beats several deep learning architectures and is only outperformed by the same algorithm \cite{landrieu_large-scale_2017}. We can notice that our results exceed Hackel et al. results \cite{hackel_fast_2016} by a large margin, consolidating the conclusion in Section 4.1.

\begin{table*}[t]
\centering
\begin{footnotesize}
\caption{$\mathsf{IoU}$ per class on Paris-Lille-3D dataset (Fold: Paris)}
\vspace{2ex}
\label{Table_Paris-Lille-3D} 
\begin{tabular}{ c | c  c  c  c  c  c  c  c  c | c }
Method  & ground & building & signage & bollard & trash cans & barriers & pedestrians & cars & vegetation  & mean \Bstrut\\
\hline
RF\_MSSF (ours) &  $99.1$ & $90.5$ & $66.4$ & $62.6$ & $5.8$ & $52.1$ & $5.7$ & $86.2$ & $84.7$ & $61.5$ \Tstrut\\
\end{tabular}
\end{footnotesize}
\end{table*}

\begin{figure*}[b]
    \centering
    \includegraphics[width=0.9\textwidth, keepaspectratio=true]{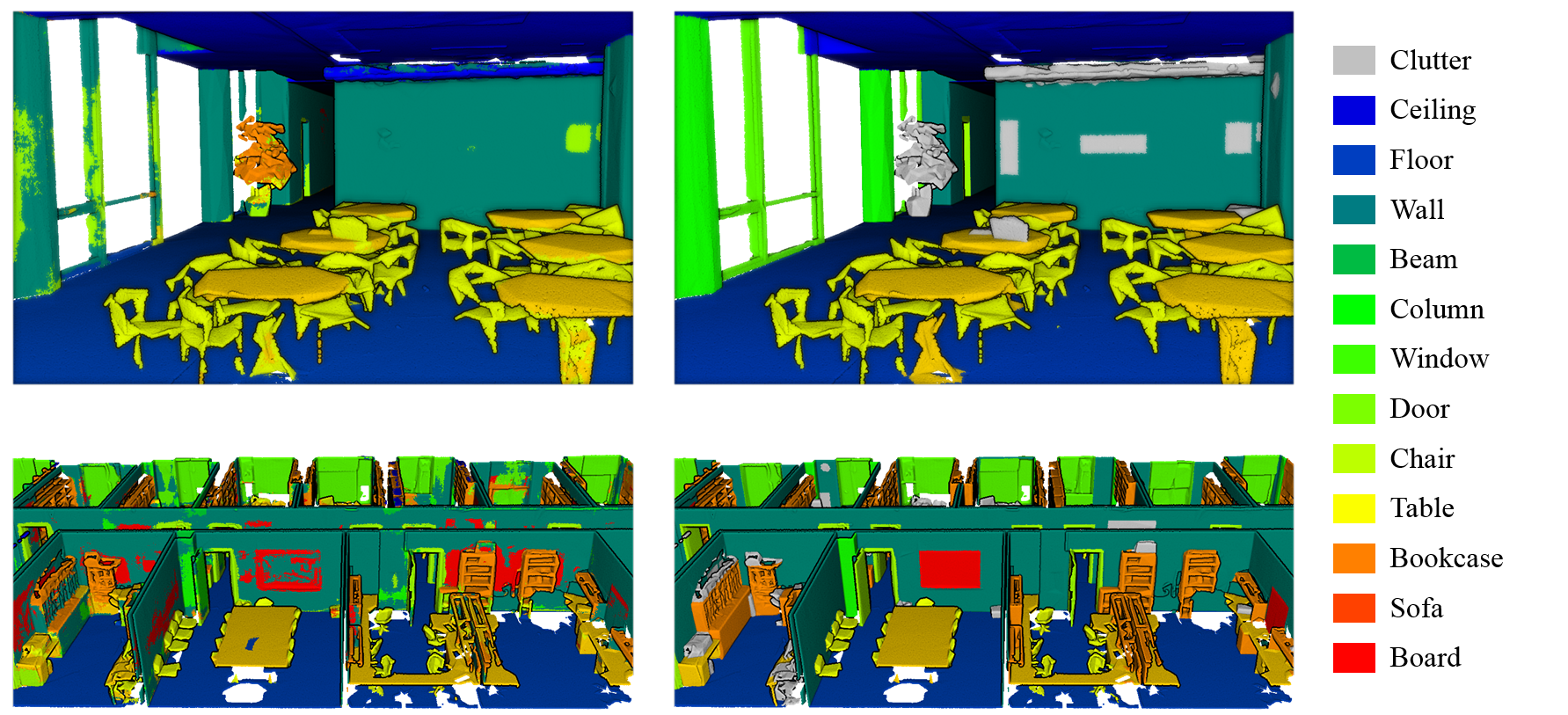}
    \caption{Examples of classified scenes in S3DIS dataset (left) with groundtruth (right)}
    \label{fig_Indoor}
\end{figure*}

Paris-Lille-3D \cite{roynard2017paris} is a recent dataset that was acquired with a Mobile Laser Scanning system in two cities in France: Lille and Paris. Overall, the scans contain more than 140 million points on $2\,km$ of streets, covering a $55000\,m^2$ area, which is much bigger than other mobile mapping datasets like Rue Madame and Rue Cassette. This dataset, fully annotated by hand, comprises 50 classes unequally distributed in three scenes Lille1, Lille2, and Paris. Following the authors' guideline, we designed 10 coarser classes defining meaningful groups: \textit{Unclassified}, \textit{Ground}, \textit{Building}, \textit{Signage}, \textit{Bollard}, \textit{Trash cans}, \textit{Barriers}, \textit{Pedestrians}, \textit{Cars}, and \textit{Vegetation}. We provide an "XML" file in supplementary materials, which maps original classes to our coarse classes. Among our ten classes, the first one \textit{Unclassified} will be ignored during training and test. We choose to train our classifier on the two scenes Lille1 and Lille2 and to use Paris as the test fold. This dataset does not include colors, so we only use our first set of features and choose the parameters used in the other outdoor environments: $S=8$, $r_0=0.1m$, $\varphi=2$, and $\rho=5$. Our results are shown in Table \ref{Table_Paris-Lille-3D}. Although this dataset is recent and does not have any other baseline result for now, we find it very interesting because of its cross-city split. We see that our classifier can transfer knowledge from one city to another and is particularly efficient on buildings. This is remarkable given that Lille and Paris architectural styles are very different.

\begin{figure*}[t]
    \centering
    \includegraphics[width=0.9\textwidth, keepaspectratio=true]{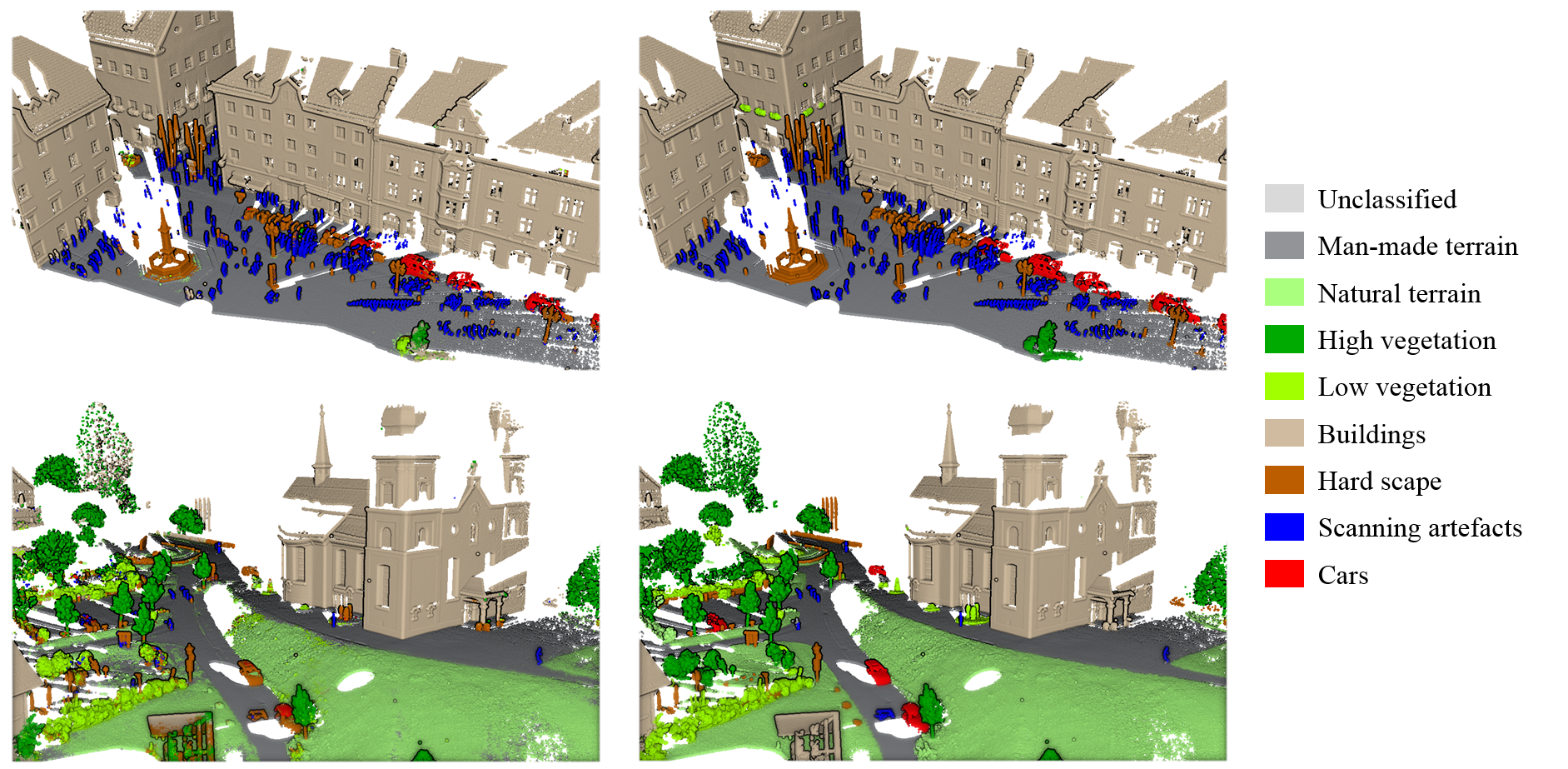}
    \caption{Examples of classified scenes in Semantic3D.net dataset (left) with groundtruth (right)}
    \label{fig_Semantic}
\end{figure*}

\begin{figure*}[t]
    \centering
    \includegraphics[width=0.9\textwidth, keepaspectratio=true]{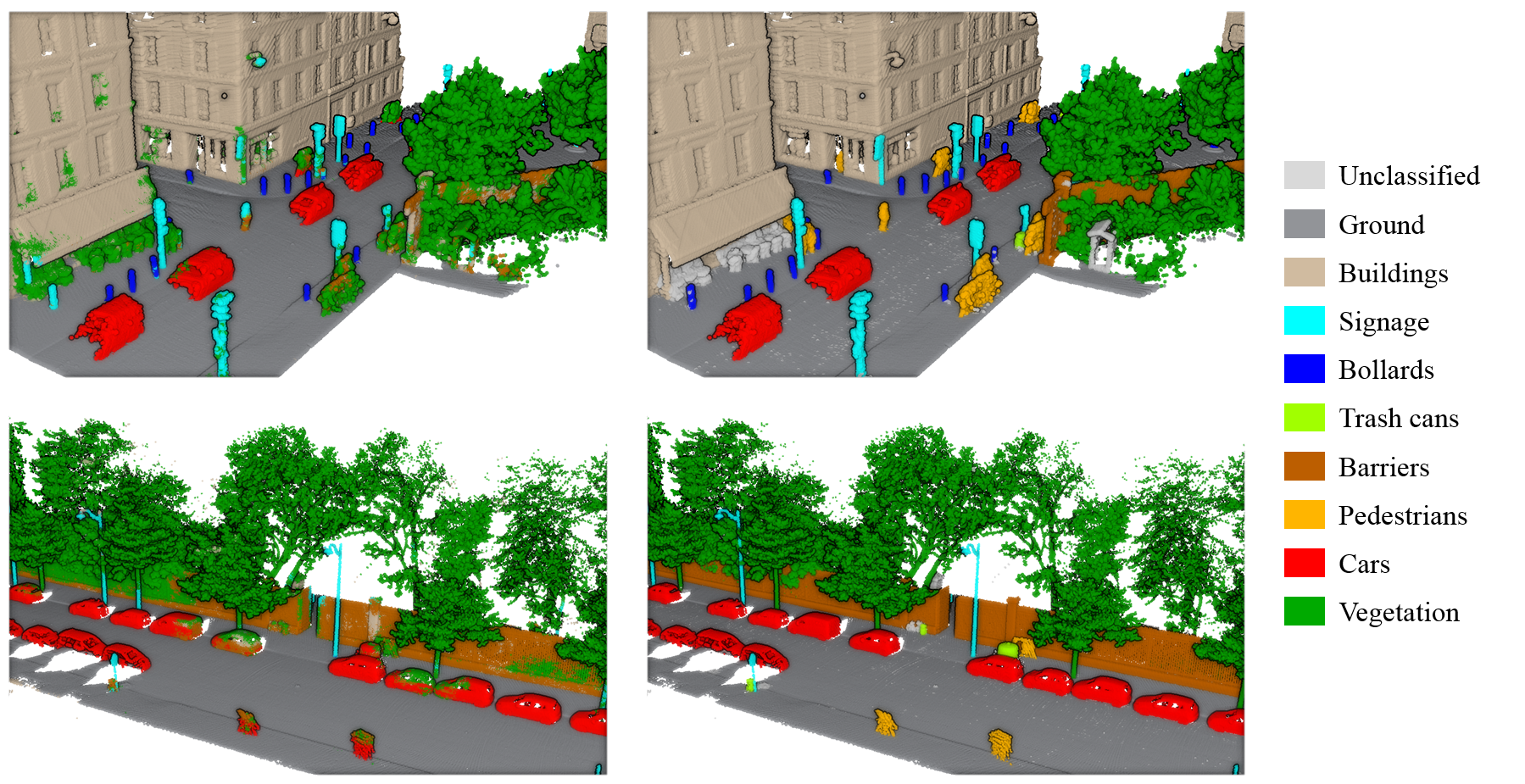}
    \caption{Examples of classified scenes in Paris-Lille-3D dataset (left) with groundtruth (right)}
    \label{fig_Paris}
\end{figure*}

Figures \ref{fig_Indoor}, \ref{fig_Semantic}, and \ref{fig_Paris} show some examples of classified scenes. First, we can notice that the classification has no object coherence as some unstructured patches appear, for example on the columns in Figure \ref{fig_Indoor} or on the facades in Figure \ref{fig_Paris}. This highlights the particularity of our method to focus on points independently, not using any segmentation scheme. Another very interesting pattern appears on the second scene in Figure \ref{fig_Paris}: when a car is close to a tree, it is misclassified and we can actually see the influence area of the tree on the car. We can assume that the classifier relies more on the large scales to distinguish those two particular classes.

Overall, our classification algorithm ranks among the best approaches, beating nearly every other elaborate method apart from Superpoint Graphs \cite{landrieu_large-scale_2017} on these datasets. However, this has to be considered in light of the fact that we do not use any segmentation or regularization process and only focus on the descriptive power of our features. We proved that our features beat state-of-the-art features in terms of classification performances, and that they could, alone, compete with complex classification schemes, including deep learning methods.

\subsection{Density parameter influence}

\begin{figure}[t]
    \centering
    \includegraphics[width=\columnwidth, keepaspectratio=true]{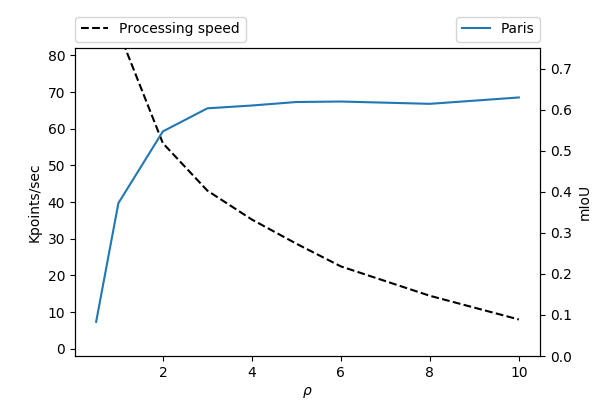}
    \centering
    \caption{Influence of the parameter $\rho$ on classification performances and computation speed on Paris-Lille-3D dataset}
    \label{fig_density}
\end{figure}

We eventually evaluate the influence of the parameter $\rho$ in our classification method. As a reminder, this parameter controls the number of subsampled points that a neighborhood can contain. A high value means better features but slower computations. In this experiment, we chose to use Paris-Lille-3D for two reasons. First, we want to focus on the 3D descriptors and, thus, do not need color information. Then, the results generalize well because they are cross-city, tested on Paris after being trained on Lille. With the parameters previously used on this dataset, we compute average  $\mathsf{IoU}$ scores across all classes for different values of $\rho$. Figure \ref{fig_density} shows the evolution of the results along with the features computation speed for every split of the dataset. We can note that average  $\mathsf{IoU}$ scores rise quickly up to $\rho=3$ and do not increase a lot for higher values of $\rho$. Depending on the application, one can choose to optimize the results or the processing speed with this parameter. Although our performances could be slightly increased with a higher $\rho$ value, we chose to keep $\rho=5$ in our work because it is a trade-off between performance and computation speed.

\section{Conclusion}

This paper presents a 3D point cloud semantic classification approach articulated around new multiscale features. The use of spherical neighborhoods instead of KNN increases the discriminating power of our features, leading to better performances than state-of-the-art features in the same experimental conditions. We also showed that the performances of our algorithm are consistent on three datasets acquired with different technologies in different environments. Eventually, we proved that our approach outperforms recent and complex classification schemes, including deep learning methods, on large scale datasets. Deep learning is becoming the standard for several classification tasks, but there is room for improvements with handcrafted methods. Furthermore, the ideas that come up from such methods, like our new multiscale neighborhood definition, could benefit other frameworks including deep learning.

\addtolength{\textheight}{-9cm}   


{\small
\bibliographystyle{ieee}
\bibliography{egbib}
}

\end{document}